\documentclass{article} 
\usepackage{iclr2020_conference,times}

\usepackage{amsmath,amsfonts,bm}

\def\eqref#1{equation~\ref{#1}}

\def\1{\bm{1}}

\DeclareMathAlphabet{\mathsfit}{\encodingdefault}{\sfdefault}{m}{sl}
\SetMathAlphabet{\mathsfit}{bold}{\encodingdefault}{\sfdefault}{bx}{n}

\usepackage{hyperref}
\usepackage{url}

\title{Neurals Networks for Projecting Named Entities from English to Ewondo}

\author{Michael Franklin Mbouopda, Paulin Melatagia Yonta* \& Guy Stéphane B. Fedim Lombo \\
Department of Computer Science \\
University of Yaounde 1 \\
Yaounde, Cameroon \\
*UMMISCO UMI 209, IRD/SU, France\\
\texttt{\{mf.mbouopda,paulinyonta,stephanefedim\}@gmail.com}
}

\iclrfinalcopy 
\begin{document}

\maketitle

\begin{abstract}
Named entity recognition is an important task in natural language processing. It is very well studied for rich language, but still under explored for low-resource languages. The main reason is that the existing techniques required a lot of annotated data to reach good performance. Recently, a new distributional representation of words has been proposed to project named entities from a rich language to a low-resource one. This representation has been coupled to a neural network in order to project named entities from English to Ewondo, a Bantu language spoken in Cameroon. Although the proposed method reached appreciable results, the size of the used neural network was too large compared to the size of the dataset. Furthermore the impact of the model parameters has not been studied. In this paper, we show experimentally that the same results can be obtained using a smaller neural network. We also emphasize the parameters that are highly correlated to the network performance. This work is a step forward to build a reliable and robust network architecture for named entity projection in low resource languages.
\end{abstract}

\section{Background}

Natural language processing (NLP) for low-resource languages is still an open challenge. Since the existing NLP models extremely rely on the amount of available data, they are not applicable to African languages. Given a parallel corpus in which one side is already annotated, annotation projection is any technique that can annotates the other side by projecting what has been learned on the annotated side \citep{yarowsky2001}. Part of speech (POS) tags have been effectively projected from French to German, Greek and Spanish \citep{zennaki2015}. The idea was to use a representation space in which a word and its translations will be very likely to have similar representations. The author proposed a binary phrase-based representation. Each word in each side is represented by a binary vector of length the number of phrases in the side of the word. The $i^{th}$ component of that vector is a $1$ if the word appeared in the $i^{th}$ phrase, otherwise it is a $0$. This representation works well for projecting POS tag, but failed to project named entity tags \citep{mbouopda2019}. To fix this limitation of the representation, \cite{mbouopda2019} suggests to take the frequency of words into consideration after observing that named entities are very less frequent in text than other expressions. From this hypothesis, \cite{mbouopda2019} has proposed a distributional representation that is more suitable for named entity projection and coupled it with a neural network to effectively project named entities from English to Ewondo. The proposed representation is based on the notion of \textit{inverse term-frequency} used in the field of information retrieval and is able to take the frequency of the word into consideration. Particularly, the representation of a word according to \cite{mbouopda2019} can be obtained from its representation regarding \cite{zennaki2015} by replacing ones by a normalized inverse of the term frequency of the word. Hence the obtained representation is no more a binary vector, but a vector of real numbers. \cite{mbouopda2019} shown that this representation is far better than the one from \cite{zennaki2015} for the projection of named entity tags.

\section{Contribution}

\cite{mbouopda2019} focused on designing a better representation for the projection of named entity tags and study the influence of the model parameter was not part of its work. He  used a neural network of two layers to successfully project named entity tags from English to Ewondo using a new representation. The first layer contained $640$ neurons while the second one contained $160$. The influence of the hyper-parameters on the projection accuracy has not been studied. Furthermore, regarding the little size of the English-Ewondo parallel corpus they used, the number of neurons in this neural network can be reduced. In fact, before \cite{mbouopda2019}, no annotated corpus exists in Ewondo, therefore they manually created a parallel English-Ewondo corpus composed of a vocabulary of $912$ words for the English side and $1029$ words for the Ewondo side. Therefore, our contribution is the assessment of the influence of the model parameters on the projection results and by the way, come out with a more optimal parameters than the ones used by \cite{mbouopda2019}.

The protocol we used to assess hyper-parameter is the following:
\begin{enumerate}
	\item Set the parameters
	\item Train the neural network using k-fold cross validation 
	\item Record the projection precision, recall and F1-score
	\item Repeat the previous steps for a $r$ iterations
	\item Average and record the precision, recall and F1-score mean and standard deviation over the $r$ iterations
	\item Restart from the first step
\end{enumerate}

For now, the parameters we try to assess at the first step are the number of neurons per layer, the number of folds, the number of iterations and the number of training epochs. Repeating the three first steps $r$ times and recording the mean and standard deviation help to check if the network is robust to random initialization of the weights.

We have performed a grid search to find optimal parameters and record the projection results for each set of parameters during the grid search. The number of neurons in each layer varies from $25$ to $1024$, We have varied the number of iteration $r$ and the number of epochs from $10$ to $30$. The number of folds varies from $5$ to $50$.

\section{Results}
The optimal configuration and the results it gives are shown in Table~\ref{tab:results}.

\begin{table}[ht]
	\caption{English to Ewondo projection result summary}
	\label{tab:results}
	\begin{center}
		\begin{tabular}{r|cccccccc}
			\multicolumn{1}{r|}{\bf id} & \multicolumn{1}{c}{\bf h1-size} & \multicolumn{1}{c}{\bf h2-size} & \multicolumn{1}{c}{\bf Epoch} & \multicolumn{1}{c}{\bf $k$} & \multicolumn{1}{c}{\bf $r$} & \multicolumn{1}{c}{\bf Precision} & \multicolumn{1}{c}{\bf Recall} & \multicolumn{1}{c}{\bf F1-score}\\
			\hline \\
			$ \#1 (baseline) $ & $ 640 $ & $ 160 $ & $ 10 $ & $ 10 $ & $ 10 $ & $ 78.68 \pm 1.25 $ & $ 64.95 \pm 2.02 $ & $ 70.52 \pm 1.53 $ \\
			$ \#2 $ & 50 & 35 & 10 & 50 & 20 & $66.87 \pm 2.74$ & $52.88 \pm 1.89$ & $59.33 \pm 1.29$ \\
			$ \#3 $ & 100 & 60 & 10 & 50 & 20 & $ 74.45\pm2.29 $ & $ 63.75\pm1.26 $ & $ 68.41\pm1.20 $ \\
			$ \textbf{\#4} $ & \textbf{300} & \textbf{225} & \textbf{10} & \textbf{10} & \textbf{10} & $\mathbf{76.34\pm2.46}$ & $\mathbf{63.64\pm2.87}$ & $\mathbf{69.22\pm1.83}$ \\
			$ \#5 $ & 400 & 300 & 10 & 10 & 10 & $78.66\pm2.01$ & $63.26\pm2.51$ & $69.01\pm1.87$ \\
			$ \#6 $ & 1024 & 840 & 15 & 20 & 10 & $76.58\pm1.32$ & $68.23\pm1.88$ & $72.36\pm1.31$ \\
		\end{tabular}
	\end{center}
\end{table}

The first line in Table~\ref{tab:results} is the network configuration used by \cite{mbouopda2019} and served here as the baseline.

We have observed that when the number of neurons per layer is small (less than $600$ for at least one layer), the number of folds is highly correlated to the network performance compared the number of iterations and the number of epochs. In this case, the higher the number of folds, the better the performances. The observation exhibit a good characteristic of the model. A larger number of fold means more data to train the model on. Since the projection performance in getting better with the number of fold, we conclude that the model is very likely to perform a better projection when trained on a larger dataset.

We can observed that the model appears to be very robust. In fact, slightly decrease or increase the number of neurons per layer do not significantly change the model performance. Even swapping the layers does not change the model performances significantly.

When the number of neurons per layer is at least $100$, and the quotient of the layer sizes is at around $0.8$, the model can get performance similar the baseline. Fore instance, the $4^{th}$ line of Table~\ref{tab:results} shows a neural network which produces similar results as the baseline by using $300$ and $225$ neurons on the first and second layers respectively.

Finally, when the number of neurons per layer is around $1000$, the neural network tends to produces performances that are slightly better than the base line. This is illustrated in Table~\ref{tab:results} by the last line.

\bibliography{iclr2020_conference}
\bibliographystyle{iclr2020_conference}

\end{document}